\pdfoutput=1

\documentclass[11pt]{article}

\usepackage[preprint]{acl}
\usepackage{times}
\usepackage{latexsym}
\usepackage{tcolorbox}
\usepackage[T1]{fontenc}

\usepackage[utf8]{inputenc}

\usepackage{microtype}
\usepackage{float}
\usepackage{inconsolata}

\usepackage{graphicx}
\usepackage{booktabs}
\usepackage{array} 
\usepackage{graphicx} 

\title{\raisebox{-0.3\height}{\includegraphics[width=1cm]{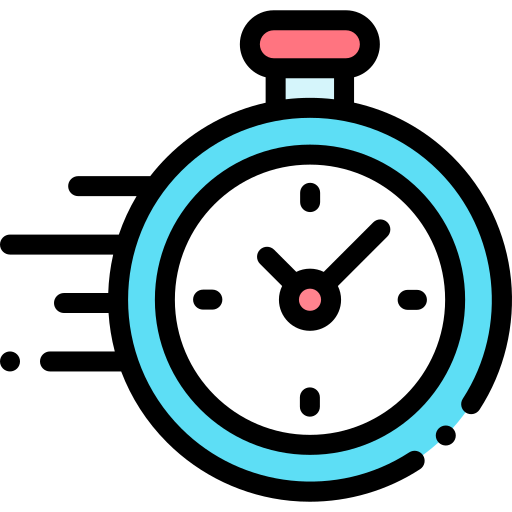}} 
    TimeCausality: Evaluating the Causal Ability in Time Dimension for Vision Language Models
}

\author{
Zeqing Wang$^*$\textsuperscript{1}
\
Shiyuan Zhang$^*$\textsuperscript{2}
\
Chengpei Tang\textsuperscript{1}
\
Keze Wang\textsuperscript{1}
\\
\textsuperscript{1}Sun Yat-sen University \qquad  \textsuperscript{2}University of Illinois Urbana-Champaign  \\
\tt\small wangzq73@mail2.sysu.edu.cn, \tt\small sz54@illinois.edu, \tt\small Tchengp@mail.sysu.edu.cn, \tt\small kezewang@gmail.com
}

\definecolor{darkpurple}{rgb}{0.31, 0.18, 0.43}

\begin{document}
\maketitle
\def\thefootnote{*}\footnotetext{These authors contributed equally to this work}
\begin{abstract}
Reasoning about temporal causality, particularly irreversible transformations of objects governed by real-world knowledge (e.g., fruit decay and human aging), is a fundamental aspect of human visual understanding. Unlike temporal perception based on simple event sequences, this form of reasoning requires a deeper comprehension of how object states change over time. Although the current powerful Vision-Language Models (VLMs) have demonstrated impressive performance on a wide range of downstream tasks, their capacity to reason about temporal causality remains underexplored. To address this gap, we introduce \textbf{TimeCausality}, a novel benchmark specifically designed to evaluate the causal reasoning ability of VLMs in the temporal dimension. Based on our TimeCausality, we find that while the current SOTA open-source VLMs have achieved performance levels comparable to closed-source models like GPT-4o on various standard visual question answering tasks, they fall significantly behind on our benchmark compared with their closed-source competitors. Furthermore, even GPT-4o exhibits a marked drop in performance on TimeCausality compared to its results on other tasks. These findings underscore the critical need to incorporate temporal causality into the evaluation and development of VLMs, and they highlight an important challenge for the open-source VLM community moving forward. Code and Data are available at \href{https://github.com/Zeqing-Wang/TimeCausality }{TimeCausality}. 
\end{abstract}

\section{Introduction}
\begin{figure}[!t]
  \centering
   \includegraphics[width=1\linewidth]{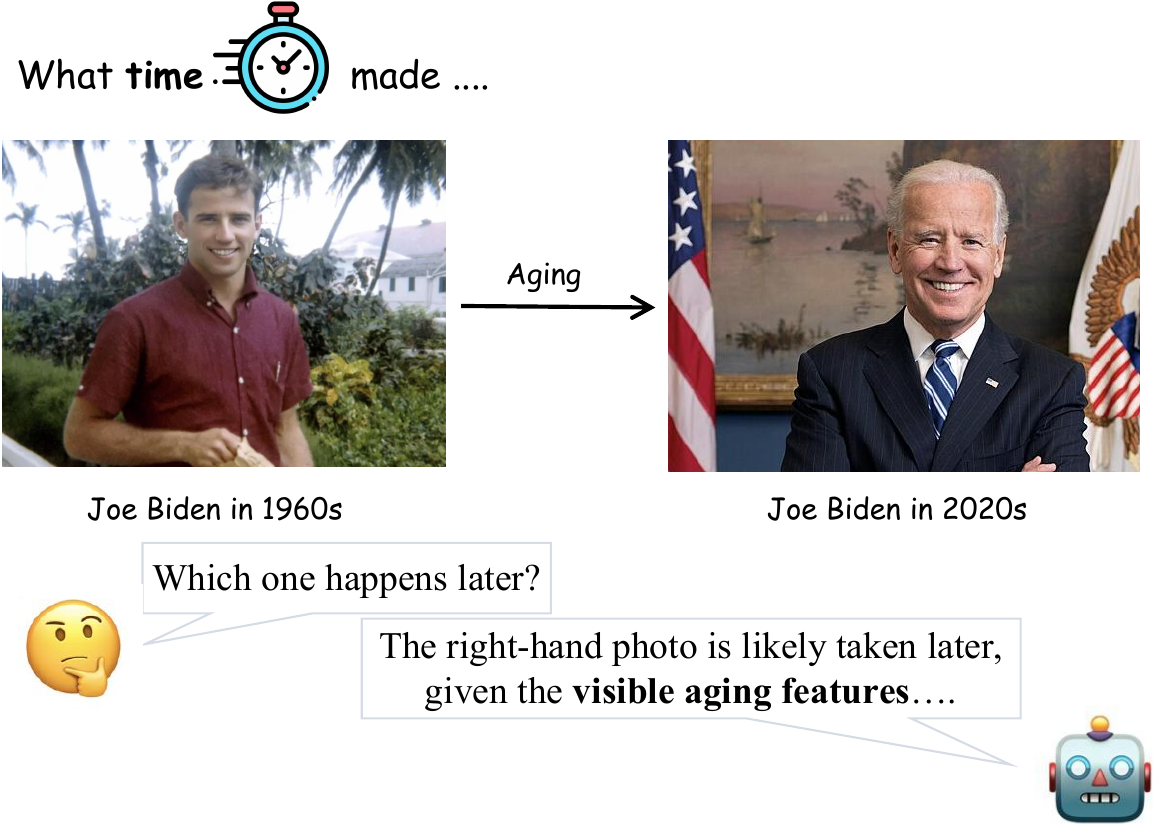}
   \caption{Illustration of our motivation. The content of an image will change over time. For example, a person will age over time, and this change is a time-dependent and irreversible process. Our goal is to evaluate VLM's ability to understand such a causal relationship.}
   \label{fig:motivation}
   \vspace{-15pt}
\end{figure}
Vision-Language Models (VLMs), such as the closed-source like GPT-4  \cite{openai2023gpt4v, openai2024gpt4ocard}, Gemini \cite{team2023gemini, team2024gemini} series, or the open-source like InternVL series \cite{chen2024internvl, luo2024mono, chen2024far}, and QwenVL series \cite{bai2023qwen, Qwen2.5-VL, Qwen2-VL}, have achieved impressive performance across a range of multimodal tasks in recent years, driven by large-scale training data and substantial model capacities. By effectively interpreting visual inputs and generating corresponding textual responses, these models have become powerful tools for a wide range of downstream applications \cite{zhang2024vision, li2023llavamed,xu2024pllava}. This progress has fueled growing interest in leveraging VLMs for tasks beyond surface-level understanding \cite{vqa, vqav2, AOKVQA}, particularly those requiring complex reasoning \cite{zhan2024anygpt,scienceQA,liu2024tempcompass,phan2025humanitysexam}. In response, a number of benchmarks have been introduced to evaluate the reasoning capabilities of VLMs across domains such as mathematics \cite{zou2024dynamath, sun2024mm}, spatial understanding \cite{liao2024reasoning}, and scientific reasoning \cite{lu2022learn}. However, many of these benchmarks are grounded in natural sciences or consist of relatively straightforward visual question answering (VQA) tasks, which often emphasize the extraction of directly observable information rather than the modeling of deeper understanding.

Furthermore, they often overlook the deeper potential of images as representations of real-world phenomena. An image is not merely a collection of pixels—it frequently encodes implicit prior knowledge, including causal, temporal, and commonsense relationships that extend beyond what is immediately observable. As shown in Figure \ref{fig:motivation}, the content of a picture is prone to change over time, which is an irreversible process caused by temporal causal relationships in nature. Some recent works \cite{naturalbench, liu2024tempcompass, li2023vitatecs, li2023seed, ning2023video, li2024mvbench} have introduced benchmarks targeting temporal aspects such as action, speed, direction, attribute changes, and event ordering. However, these evaluations neglect causal reasoning grounded in human knowledge and the nature of processing within the visual information, such as prolonged exposure of metal to the environment causes corrosion, and human skin develops irreversible wrinkles with aging.

This gap reveals a limitation in current VLM evaluation: the absence of benchmarks targeting temporal-causal reasoning based on real-world knowledge. To bridge this gap and promote progress in temporal causal understanding for VLMs, we identify two key questions that should be answered: (1) To what extent do current VLMs understand real-world temporal causality? (2)What are the root causes of model failure on such tasks—are the limitations due to a lack of reasoning ability, or an inability to identify causally relevant cues?

To this end, in this paper, we introduce a novel benchmark, \textbf{TimeCausality}, designed to assess the VLMs' capacity of causal reasoning in the temporal dimension. Our benchmark centers on five distinct categories of causal factors: Physical Change (PC), Chemical Change (CC), Natural Phenomenon (NP), Environmental Modification (EM), and Artificial Processing (AP). In contrast to previous benchmarks, our TimeCausality focuses explicitly on causal reasoning over time within an image pair, challenging models to infer commonsense knowledge that extends beyond surface-level visual cues and provides rationales.

Our main contributions can be summarized as follows: (i) We introduce \textbf{TimeCausality}, the first benchmark specifically targeting real-world temporal causal reasoning, constructed with high-quality, human-verified image pairs. (ii) We propose a three-aspect evaluation to systematically assess whether VLMs can comprehend causal relationships grounded in temporal progression. (iii) Through extensive evaluations across a diverse set of VLMs with varying scales and architectures, we highlight the significant challenges these models face in understanding complex temporal-causal relationships.

\section{Related Work}

\subsection{Vision-Language Models}
VLMs, built upon extensive pretraining datasets and integrating the capabilities of Large Language Models (LLMs)~\cite{gpt3,llama,llama2,zeng2022glm,team2024gemma,jiang2023mistral,bi2024deepseek,flant5} with advanced vision encoders~\cite{vit, blip2, he2024bunny, tong2024cambrian1}, have demonstrated exceptional performance across a wide range of visual tasks. Closed-source VLMs, such as GPT-4 \cite{openai2023gpt4v,gpt4o}, Claude 3.5 \cite{anthropic2024claude3} and Gemini \cite{team2023gemini, team2024gemini}, excel in a wide range of perception and reasoning tasks. Meanwhile, the community has developed several open-source VLMs, including InternVL \cite{chen2024internvl}, Qwen2-VL  \cite{bai2023qwen}, Llama3.2-Vision \cite{dubey2024llama}, GLM4 \cite{glm2024chatglm}, Phi3.5-Vision \cite{abdin2024phi}, etc., which demonstrate competitive performance on specific tasks. However, our study reveals that while open-source VLMs often match or even surpass closed-source counterparts on many benchmarks, there remains a substantial performance gap on our proposed TimeCasualty, which highlights a significant challenge for the community to address.

\subsection{Benchmarks for VLM }
With the rapid development of VLMs, an increasing number of studies have aimed to evaluate their capabilities from diverse perspectives. Unlike traditional VQA tasks, which primarily assess a model’s visual understanding \cite{vqa,vqav2,gqa} or outside knowledge \cite{AOKVQA}, recent benchmarks have begun to focus on the deeper reasoning abilities of VLMs. For instance, POPE \cite{Li-hallucination-2023} evaluates the extent of hallucination in VLMs, while benchmarks such as MMMU \cite{yue2023mmmu} and ScienceQA \cite{scienceQA} assess scientific reasoning skills. More recently, benchmarks like Impossible Video \cite{bai2025impossible} and TempCompass \cite{liu2024tempcompass} have targeted VLMs’ understanding of commonsense knowledge. However, existing benchmarks suffer from two main limitations: (1) they often lack explicit commonsense rationales, and (2) the commonsense questions included are typically straightforward, posing little challenge for increasingly advanced VLMs. In contrast, our proposed TimeCasualty evaluates VLMs' understanding of real-world commonsense from a temporal perspective. It provides high-quality rationales and features challenging questions that push the limits of current VLMs’ temporal commonsense reasoning.

\section{TimeCasualty Benchmark}
Temporal causality refers to the process by which a scene or object undergoes state changes over time due to underlying causal factors in the world. Humans naturally infer the temporal progression of object states by leveraging prior knowledge about irreversible real-world processes (e.g., human aging). Our TimeCausality is designed to evaluate this capacity in VLMs, by assessing their ability to infer and reason about such temporally grounded causal relationships from visual information.

\subsection{Time Casualty Types}
To systematically evaluate temporal causal reasoning for VLMs, we categorize temporal-induced transformations into five distinct types based on their underlying causal mechanisms and the domain knowledge required to interpret their progression. As illustrated in Figure~\ref{fig:casual_types}, the five categories are defined as follows:

\begin{figure*}[!t]

\centering

\includegraphics[width=0.8\linewidth]{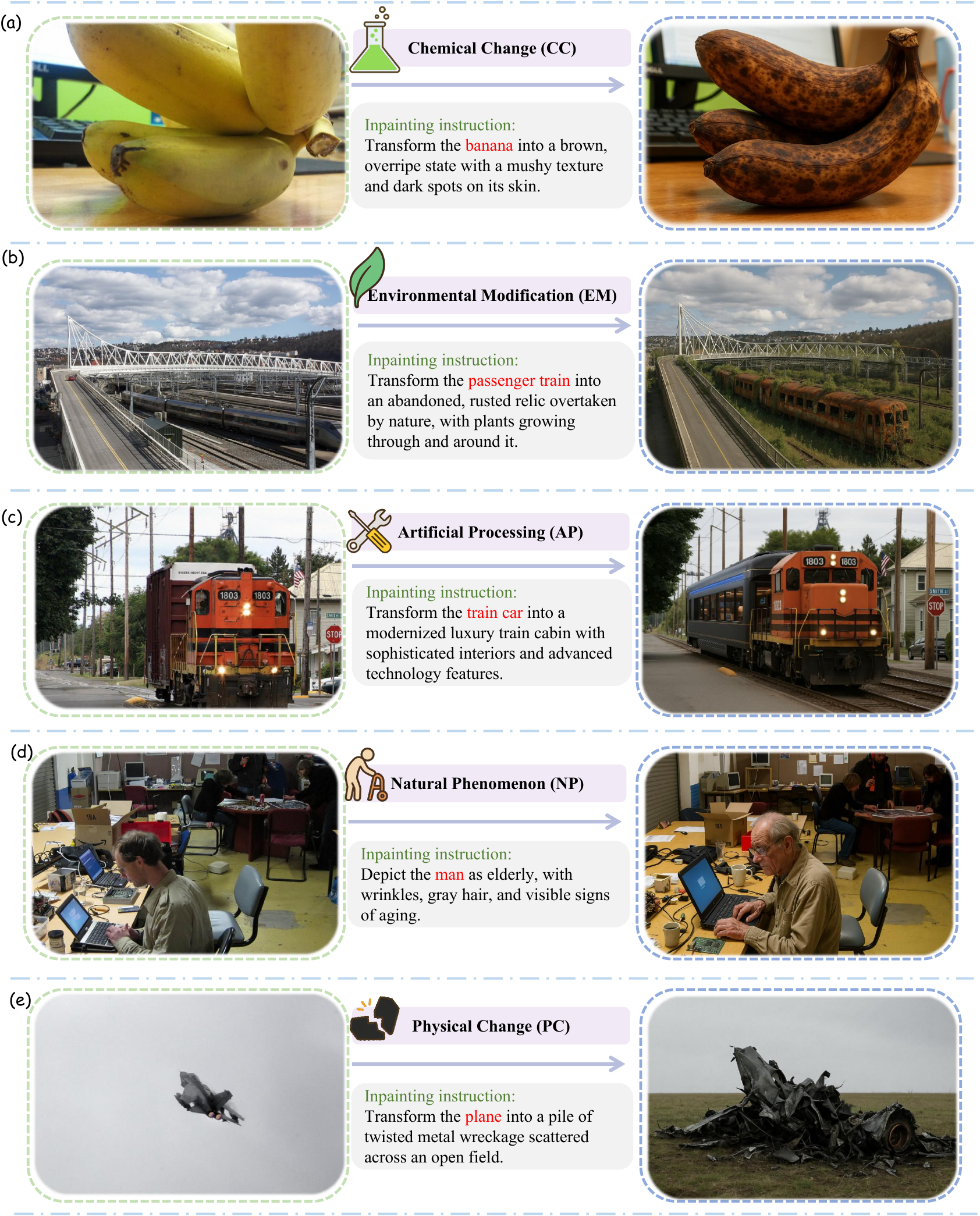}

\caption{Causal types in our \textbf{TimeCausality} benchmark. We identify five dimensions of temporal causality: Artificial Processing (AP), Chemical Change (CC), Natural Phenomenon (NP), Environmental Modification (EM), and Physical Change (PC). The image depicting the causal result is generated by modifying the original using an inpainting model. The object marked in \textcolor{red}{RED} denotes the primary target of causality.}
\vspace{-0.5cm}
\label{fig:casual_types}

\end{figure*}

\textbf{Chemical Change (CC)}: This category involves irreversible chemical transformations, such as food spoilage or oxidation. For example, in Figure~\ref{fig:casual_types} (a), a banana becomes overripe over time.

\textbf{Environmental Modification (EM)}: This includes changes in the environment induced by natural forces, such as deforestation or urbanization. In Figure~\ref{fig:casual_types} (b), the train slowly rusts and the plants around it grow through because it is abandoned. 

\textbf{Artificial Processing (AP)}: This type refers to human-induced transformations, such as manufacturing or construction. For instance, as shown in Figure~\ref{fig:casual_types} (c), a train car may be modernized through human intervention. 

\textbf{Natural Phenomenon (NP)}: This includes naturally occurring processes such as plant growth, seasonal variation, or aging. For example, Figure~\ref{fig:casual_types} (d) illustrates the irreversible aging of a person. 

\textbf{Physical Change (PC)}: This refers to transformations in physical states or structures, such as deformation, collapse, or motion. As depicted in Figure~\ref{fig:casual_types} (e), an aircraft is irreversibly damaged due to external impact. 

It is worth noting that while all five causal types are governed by the passage of time, the direction and nature of these changes can vary significantly depending on the object involved. For instance, in Artificial Processing (AP), time often reflects constructive human intervention, transforming raw materials into refined products. In contrast, Environmental Modification (EM) typically involves degradative or erosive effects, such as natural decay or corrosion. Although both are time-driven, they produce opposite outcomes. This contextual variability poses a unique challenge for VLMs, as models must not only recognize that a temporal change has occurred, but also infer its directionality and underlying causal mechanism based on prior knowledge about the object and environment.

\subsection{Benchmark Structure}
\begin{figure*}[!t]
  \centering
   \includegraphics[width=1\linewidth]{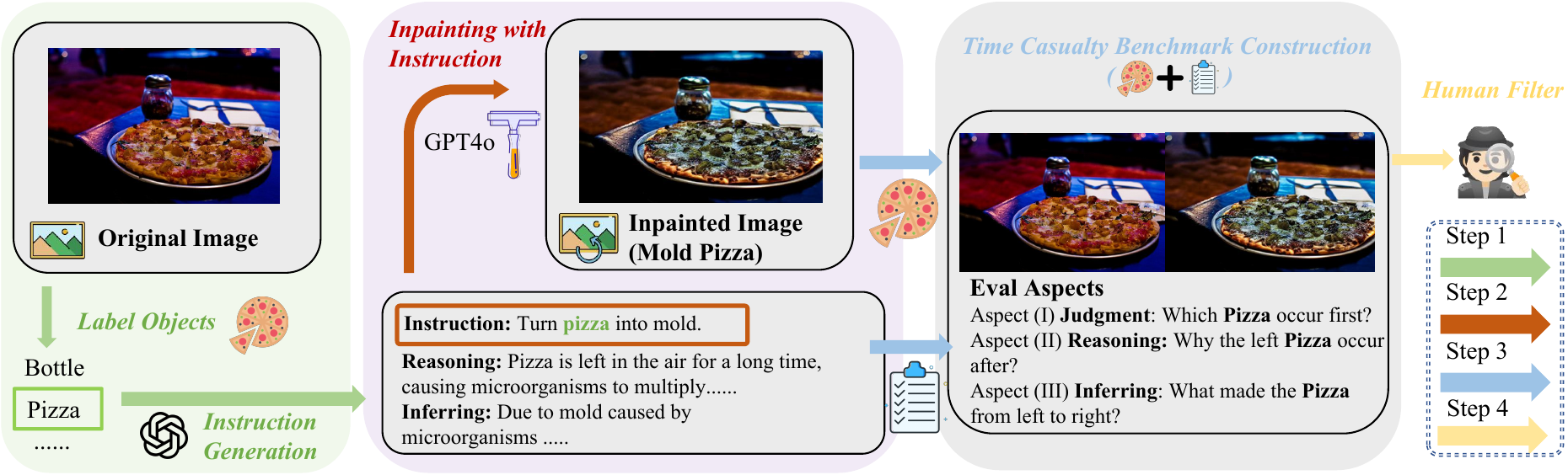}
   \caption{Illustration of TimeCausality generation pipeline. First, we use a grounding model with an auto-labeled model to detect and label key objects in the original image. Based on the detected objects, an LLM generates possible irreversible transformations that the object could undergo over time and generates corresponding modification instructions along with reasoning explanations (e.g., turning a pizza into a moldy state). Next, we apply the powerful GPT-4o as the inpainting model to modify the original image according to the generated editing instruction, creating a temporally evolved version of the object within the image. Finally, we pair the original and modified images and incorporate the reasoning and inferring explanation from the instruction step as the ground truth to construct our TimeCausality. All data are manually checked following the instructions provided in Appendix~\ref{sec:human_verification}.}
   \label{fig:data_pipeline}
   \vspace{-0.5cm}
\end{figure*}

\subsubsection{Dataset Collection and Generation}
To construct our TimeCausality based on the above definition, we develop an automated pipeline with human verification. As shown in Figure \ref{fig:data_pipeline}, the process begins with high-quality wild images sourced from the COCO 2017 validation split \cite{lin2014microsoft}, which serves as the foundational data. To obtain fine-grained object-level information for evaluation purposes, we employ an Auto Label model \cite{ren2024grounded} to detect and identify objects within these images. Subsequently, these object labels are utilized to guide an LLM in generating detailed textual annotations, including (i) image editing instructions, (ii) transition reasoning, and (iii) inferring explanations, which will be used as the rationales of our TimeCausality.

Then, leveraging the editing instructions generated by LLM, we employ the powerful GPT-4o as the inpainting model to transform the original image into a temporally modified version. Then, to ensure the modified object is reasonable and suitable, we filter out the objects that with too small or too large an area.

Finally, to ensure the quality of the automatically processed dataset for evaluation purposes, we implement a rigorous manual verification process and conduct careful curation of high-quality samples. Given that capturing temporal causality requires substantial domain knowledge, we engage multiple annotators, each possessing at least an undergraduate degree, to manually validate the samples in the dataset. All prompts and pre-trained models utilised throughout the pipeline are shown in Appendix \ref{subsec:prompt_generation}.

It is worth noting that a critical aspect of our dataset construction is the emphasis on \textbf{irreversible} transformations. These transformations serve as a fundamental principle throughout our processing pipeline and are the core target of our benchmark: establishing clear temporal causality. This dictates that edited images must exhibit an unambiguous temporal progression from their original counterparts, thereby ensuring a robust temporal relationship between the source and modified images.

However, given that a single image contains a wealth of information, we implement specific measures in two key areas to ensure the VLM accurately comprehends the objects undergoing these irreversible transformations. (1) Dataset Construction: We endeavor to preserve the overall environmental context, isolating modifications primarily to the target object of interest (unless the environment itself was integral to the intended temporal change). This selective alteration strategy directs the VLM's focus to the specific object that has been transformed. (2) Evaluation Prompting: During evaluation, our prompts are designed to explicitly highlight the detection target, for instance, by incorporating the object's name in the prompt. Furthermore, to mitigate potential biases from the input order of images, we form image pairs by vertically concatenating them. Further details on these procedures are provided in Section \ref{sec:exp}.

\subsubsection{Data Format}
\begin{figure*}[t]
  \centering
   \includegraphics[width=\linewidth]{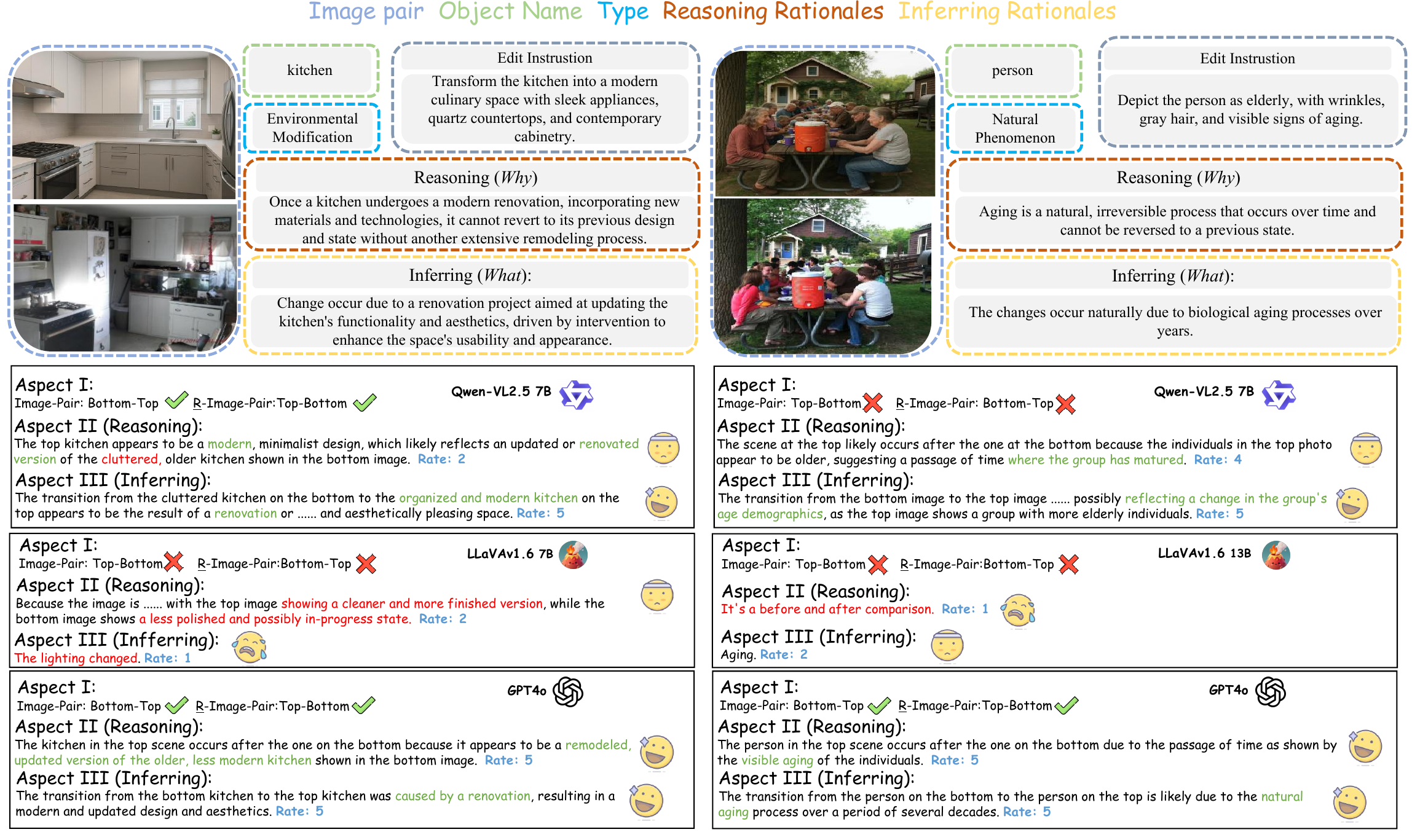}
   \caption{Data demo and case study in our TimeCausality.}
   \label{fig:data_demo}
\end{figure*}

As shown in Figure \ref{fig:data_demo}, each sample in \textbf{TimeCasualty} comprises five elements: \textbf{Image pair}, \textbf{Object Name}, \textbf{Type}, \textbf{Reasoning Rationales}, and \textbf{Inferring Rationales}. The Object Name means the object injected with casualty within the image-modified process. The \textbf{Type} specifies the category of the temporal-causal relationship, corresponding to the Figure \ref{fig:casual_types}. The \textbf{Reasoning Rationales} describes \textit{why} the irreversible casualty exists, and the \textbf{Inferring} focuses describes \textit{what} caused the temporal casualty. We also provide more samples in our TimeCausality in the Appendix \ref{sec:more_cases_in_time}.

\subsection{Evalutation}
\begin{figure}[H]
  \centering
   \includegraphics[width=0.9\linewidth]{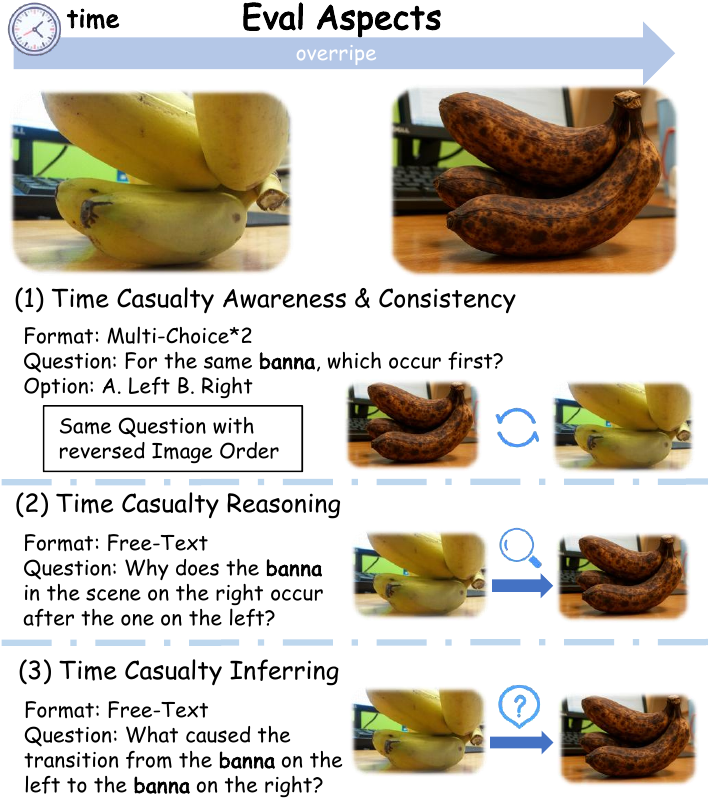}
   \caption{Evaluation aspects of our TimeCausality.}
   \label{fig:data_eval}
\end{figure}

Based on the data format, we design three unique but interconnected evaluation aspects, as shown in Figure \ref{fig:data_eval}. The details are as follows:

\paragraph{Aspect I: Casualty Awareness \& Consistency. (\textit{Which})} In this aspect, given an image pair, the VLM should be able to determine the correct temporal order. It is worth noting that the image pair contains the order nature, to avoid the basis of the evaluation, we test both image orders in this evaluation to evaluate the consistency of the model. 

\paragraph{Aspect II: Casualty Reasoning. (\textit{Why})} By asking why the latter object occurs after the prior object, this aspect aims to evaluate the model’s ability to recognize causal links and provide plausible explanations grounded in visual cues. 

\paragraph{Aspect III: Casualty Inferring. (\textit{What})} Compared with the evaluation on ``Casualty Reasoning'', this question type challenges the model to infer the cause of the transition from the previous state to the later state, which requires a deeper commonsense in the real-world and the ability in visual reasoning.

Together, these three question types provide a holistic framework to evaluate VLMs' temporal-causal reasoning abilities. Overall, the \textit{Aspect I} is in the format of multiple-choice, and \textit{Aspect II \& III} are in free-text format.

\subsection{Dataset Statistics}
\begin{figure}[!t]
  \centering
   \includegraphics[width=0.9\linewidth]{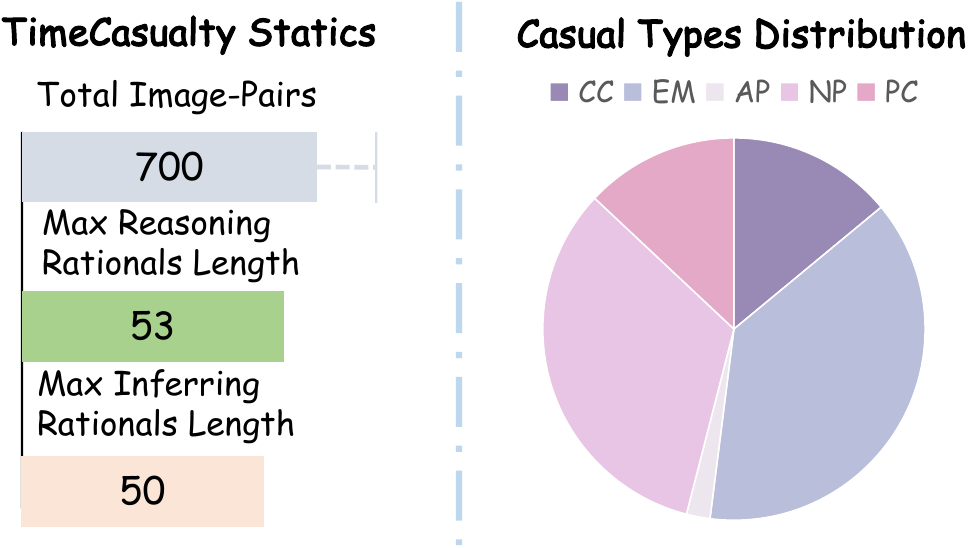}
   \caption{Data statistics in our TimeCasualty.}
   \label{fig:data_statistic}
\end{figure}

In Table \ref{fig:data_statistic}, we provide the basic information about our TimeCausality. In summary, the TimeCasuality comprised 700 image pairs, and the max rationales length of reasoning and inferring correspond to 53 and 50.

\section{Experiment}
\label{sec:exp}

\begin{table*}[t!]
  \centering
  \resizebox{0.9\textwidth}{!}{%
    \begin{tabular}{lrrrrcc}
    \toprule
    Metric & \multicolumn{4}{c}{Aspect I}  & Aspect II & Aspect III \\
    \midrule
    Model & \multicolumn{1}{l}{ACC} & \multicolumn{1}{l}{ACC-R} & \multicolumn{1}{l}{Group Score} & \multicolumn{1}{l}{F1 Score} & \multicolumn{1}{l}{Reasoning Score} & \multicolumn{1}{l}{Inferring Score} \\
    \midrule
    \multicolumn{7}{c}{\textcolor{orange}{Open-Source Vision Language VLMs}} \\
    \midrule
    Mono-InternVL-2B \cite{luo2024mono} & 62.86 & 34.57 & 20.00 & 47.67 & \multicolumn{1}{r}{0.83} & \multicolumn{1}{r}{1.34} \\
    InternVL2-4B \cite{chen2024internvl}& 51.71 & 47.14 & 24.86 & 49.40 & \multicolumn{1}{r}{0.60} & \multicolumn{1}{r}{0.67} \\
    InternVL2-8B \cite{chen2024internvl}& 62.86 & 36.00 & 22.00 & 48.50 & \multicolumn{1}{r}{0.57} & \multicolumn{1}{r}{1.02} \\
    InternVL2\_5-4B \cite{chen2024internvl}& 45.43 & 45.71 & 24.86 & 45.57 & \multicolumn{1}{r}{1.12} & \multicolumn{1}{r}{1.16} \\
    InternVL2\_5-8B \cite{chen2024internvl}& 26.57 & 78.57 & 20.86 & 49.13 & \multicolumn{1}{r}{1.59} & \multicolumn{1}{r}{1.91} \\
    Qwen2-VL-2B-Instruct \cite{Qwen2-VL}& 43.71 & 59.14 & 27.71 & 51.14 & \multicolumn{1}{r}{0.76} & \multicolumn{1}{r}{1.13} \\
    Qwen2-VL-7B-Instruct \cite{Qwen2-VL} & 21.43 & 78.29 & 16.00 & 45.45 & \multicolumn{1}{r}{0.87} & \multicolumn{1}{r}{1.31} \\
    Qwen2.5-VL-3B-Instruct \cite{Qwen2.5-VL}& 49.43 & 52.29 & 26.29 & 50.85 & \multicolumn{1}{r}{1.55} & \multicolumn{1}{r}{1.58} \\
    Qwen2.5-VL-7B-Instruct \cite{Qwen2.5-VL}& 41.14 & 71.14 & 28.00 & 55.13 & \multicolumn{1}{r}{2.23} & \multicolumn{1}{r}{2.29} \\
    Cogvlm2-llama3-chat-19B \cite{hong2024cogvlm2}& 50.86 & 22.00 & 12.86 & 35.08 & \multicolumn{1}{r}{0.62} & \multicolumn{1}{r}{1.11} \\
    Llama3.2-vision-11B \cite{grattafiori2024llama} & 79.14 & 1.43  & 1.43  & 29.67 & \multicolumn{1}{r}{1.02} & \multicolumn{1}{r}{2.38} \\
    LLava-v1.6-7B \cite{liu2024llavanext}& 66.29 & 29.43 & 19.14 & 46.02 & \multicolumn{1}{r}{0.96} & \multicolumn{1}{r}{0.72} \\
    LLava-v1.6-13B \cite{liu2024llavanext}& 50.00 & 52.86 & 25.43 & 51.42 & \multicolumn{1}{r}{1.01} & \multicolumn{1}{r}{1.24} \\
    GLM4-9B \cite{glm2024chatglm}& 59.71 & 27.14 & 14.29 & 41.89 & \multicolumn{1}{r}{1.32} & \multicolumn{1}{r}{1.76} \\
    Phi-3-vision-128k-instruct \cite{abdin2024phi}& 81.71 & 15.14 & 10.86 & 42.00 & \multicolumn{1}{r}{0.65} & \multicolumn{1}{r}{0.84} \\
    Phi-3.5-vision-instruct \cite{abdin2024phi}& 69.71 & 13.71 & 8.86  & 36.76 & \multicolumn{1}{r}{0.44} & \multicolumn{1}{r}{0.76} \\
    \midrule
    \multicolumn{7}{c}{\textcolor{blue}{Closed-Source Vision Language VLMs}} \\
    \midrule
    GPT-4o-mini \cite{openai2024gpt4ocard}& \textbf{98.57} & 1.43  & 1.43  & 34.56 & \multicolumn{1}{r}{2.04} & \multicolumn{1}{r}{2.37} \\
    GPT-4o \cite{openai2024gpt4ocard}& 70.86 & 64.86 & \textbf{43.43} & \textbf{67.83} & \multicolumn{1}{r}{\textbf{2.45}} & \multicolumn{1}{r}{\textbf{2.80}} \\
    \midrule
    \multicolumn{7}{c}{\textcolor{red}{Contrastive Learning Based VLMs}} \\
    \midrule
    clip-vit-base-patch16 \cite{radford2021learning} & 42.86 & 56.86 & 9.14  & 49.61 & \multicolumn{1}{r}{\textbackslash{}} & \multicolumn{1}{r}{\textbackslash{}} \\
    clip-vit-base-patch32 \cite{radford2021learning}& 32.29 & 60.86 & 8.29  & 45.46 & \multicolumn{1}{r}{\textbackslash{}} & \multicolumn{1}{r}{\textbackslash{}} \\
    clip-vit-large-patch14-336 \cite{radford2021learning}& 15.42 & \textbf{85.14} & 6.28  & 43.41 & \multicolumn{1}{r}{\textbackslash{}} & \multicolumn{1}{r}{\textbackslash{}} \\
    \bottomrule
    \end{tabular}%
  }

\caption{The performance of a diverse range of VLMs on our TimeCasualty, which includes \textcolor{orange}{Open-Source VLMs} (e.g., the LLaVA and InternVL series), \textcolor{blue}{Closed-Source VLMs} (e.g., GPT-4o), and \ \textcolor{red}{Contrastive Learning-based VLMs} (e.g., CLIP). The results indicate that most models exhibit performance inconsistencies, characterized by relatively high ACC or ACC-R scores but lower F1 Scores and Group Scores. In contrast, GPT-4o achieves state-of-the-art performance on our TimeCasualty. This observation suggests that while open-source models are progressively narrowing the performance gap with closed-source models on many benchmarks, their performance remains notably limited on tasks like TimeCasualty that require a strong understanding of real-world causality. \textbf{Bold font} indicates the best performance in each column. Note that ACC and ACC-R may not reflect the true performance of our TimeCausality, potentially due to position bias in the VLMs.} 
\vspace{-0.5cm}
\label{tab:test_results}
\end{table*}

\subsection{Experiment Setups}
\textbf{Model Selection.} To comprehensively evaluate the ability of time casualty reasoning in current VLMs, we evaluate different types and sizes of VLMs, including the current SOTA \textcolor{orange}{Open-Source VLM}, the powerful \textcolor{blue}{Closed-Source VLM} GPT4o and GPT4o-mini, and \textcolor{red}{Contrastive Learning-Based VLMs}. For experimental details, refer to the Appendix~\ref{sec:exp_deatails}.

\subsection{Evaluation Metrics}
Based on the evaluation aspect shown in Figure \ref{fig:data_eval},
We use the accuracy evaluation metric for Aspect I (multiple-choice format). For Aspect II \& III, in the free-text format, similar to many other methods \cite{zheng2023judging}, we use an LLM as an evaluator to rate the output from the model against the ground truth. The detailed prompts used in the evaluation are shown in Appendix \ref{sec:prompt_evaluation}.

\textbf{Evaluation in Aspect I:} In this evaluation, we first calculate the ACC, ACC-R, which means the average accuracy on the image-pairs and the reversed-image-pairs. Then, based on the two metrics, we further calculate the Group-Score and the F1-Score to evaluate the consistency.

\textbf{Evaluation In Aspect II \& III:} For \textbf{Aspect II\&III} evaluation, since the responses are in free-text format, we use the powerful Llama3 8B as the evaluator to compare the results generated by VLMs with the ground truth. Then, with a rating from 0 to 5 (the higher score represents a better semantic similarity between the output with the ground truth). We calculate the average of the ratings as the Reason Score and Infer Score.

\subsection{Main Results}
We test our TimeCausality on various types and sizes of VLMs. As shown in Table \ref{tab:test_results}, we report all six metrics for the current SOTA Open-Source VLMs (from 2B to 19B with different series) and the powerful commercial Closed-Source VLMs, and the evaluation on Aspect I for the Contrastive Learning-based VLMs (CLIP). It shows that, for most Open-Sourced VLMs, all metrics on the evaluation of Aspect I are limited, but the results on Aspect II \& III are relatively acceptable. 

For the powerful Closed-Source VLMs, the GPT-4o achieves the best performance for Aspect I, with a huge gap compared with the Open-Source VLMs. However, for Aspect II \& III, the GPT-4o does not stay ahead like Aspect I, some open-source Models, like Qwen2.5-VL-7B-Instruct, achieve similar performance to GPT-4o. 

We attribute this performance gap to several key factors: (1) The superior overall capabilities of GPT-4o allow it to outperform Open-Source VLMs across all aspects. In contrast, Open-Source VLMs—constrained by model size and the limitations of their training data—often exhibit significant biases, particularly in Aspect I. For example, Llama3.2-vision-11B tends to consistently predict a fixed temporal order, leading to inflated ACC but extremely poor ACC-R. Similarly, most Open-Source VLMs perform poorly on more stringent metrics in Aspect I, such as Group Score (correctly identifying both image pairs in a question) and F1 score. (2) However, for Aspect II and Aspect III, the prompts (as shown in Appendix \ref{sec:prompt_evaluation}) explicitly provide the correct temporal sequence, requiring models to only infer the cause behind the state change. This type of reasoning depends more on general commonsense knowledge about the real world—an area where many open-source LLMs already demonstrate reasonable competence. As a result, some open-source models, such as Qwen2.5-VL-7B-Instruct, achieve comparable performance to GPT-4o in these aspects.

Overall, the performance disparity between open and Closed-Source VLMs is most pronounced in Aspect I, which suggests that while both categories of models show some capacity for causal reasoning (i.e., Aspect II and III), Closed-Source models possess a substantial advantage in aligning visual content with textual understanding to make correct temporal judgments, a skill that is crucial for the tasks in TimeCausality and remains a challenge for Open-Source VLMs.

\subsection{Deeper Analysis}
\textbf{Case Study.}
To further analyze the specific challenges that current VLMs face with our TimeCausality, we present illustrative examples in Figure \ref{fig:data_demo}. In the first example (left), we establish temporal causality by depicting the modernization of an old kitchen. The irreversibility in this example is rooted in the common observation that an old kitchen is typically updated to a modern one, while the reverse transformation—a modern kitchen being reverted to an outdated state—is highly improbable. On this specific sample, both QwenVL2.5-7B and GPT-4o correctly assess Aspect I. In contrast, LLaVAv1.6-7B fails across all three of our defined evaluation dimensions. Notably, although QwenVL2.5-7B accurately addresses Aspect I, its reasoning process exhibits an inaccuracy. While much of its generated explanation is valid, it incorrectly cites the kitchen being `cluttered' as the reason for the established temporal order.

In the second example, temporal causality is represented by the natural phenomenon of human aging. Among the evaluated models—Qwen-VL2.5 7B, LLaVAv1.6 13B, and GPT-4o—only GPT-4o consistently demonstrated strong performance across all three Aspects. Notably, Qwen-VL2.5 7B, despite good comprehension on Aspect II and Aspect III, failed entirely on Aspect I. This outcome suggests that current VLMs struggle to make correct judgments without explicit prior information. Crucially, the temporal order of images was provided during the evaluation of Aspect II and Aspect III, but not for Aspect I. This further implies that while VLMs may possess the necessary common sense to address such problems, they find it challenging to directly apply this knowledge to the specific task to arrive at accurate conclusions. The powerful GPT-4o has a similar phoneme, we provide a case in the Appendix \ref{sec:a_sepcial_case}.

\textbf{Difference among the Casualty types.}
For a more granular analysis of how current VLMs perform on our TimeCausality, Table \ref{tab:sota_closesource} and Table \ref{tab:sota_opensource} present detailed performance metrics for SOTA Open-Source VLM (QwenVL2.5 7B) and GPT-4o, respectively. These results are further categorized by different causal types.

Specifically, GPT-4o demonstrates relatively balanced performance across all causal types, whereas QwenVL2.5 7B shows notable weaknesses in the \textbf{CC} and \textbf{NP} types. We analyze that this is because changes in these two types often occupy only a small portion of the entire image, requiring the model to effectively associate subtle visual cues with underlying causal reasoning,  which demands the model’s ability to align visual perception with temporal reasoning, which accounts for the observed performance gap. This finding is consistent with our earlier analysis in Table \ref{tab:test_results}, where Open-Source VLMs, including Qwen-VL, struggled in Aspect I due to limited visual-language associated capacity.

\begin{table}[!t]
  \centering
  \resizebox{1\linewidth}{!}{%
    \begin{tabular}{ccccccc}
    \toprule
    Model & \multicolumn{6}{c}{GPT 4o} \\
    \midrule
    Casual Types & ACC   & ACC-R & ACC Group & F1 Score & R-Score & I-Score \\
    \midrule
    CC    & 83.33 & 52.08 & 37.50 & 66.90 & 2.85  & 2.83 \\
    EM    & 56.82 & 78.03 & 40.91 & 67.05 & 3.05  & 3.86 \\
    AP    & 75.00 & 62.50 & 37.50 & 68.63 & 1.50  & 2.50 \\
    NP    & 80.34 & 57.26 & 48.72 & 68.38 & 1.39  & 1.44 \\
    PC    & 73.33 & 60.00 & 44.44 & 66.52 & 3.18  & 3.24 \\
    \bottomrule
    \end{tabular}%
    }
    \caption{Classification performance of the GPT4o, which achieves the SOTA performance among all the VLMs in our TimeCasualty. R-Score denotes the Reasoning Score, and I-Score denotes the Inferring Score.}
    \vspace{-0.5cm}
  \label{tab:sota_closesource}%
\end{table}%

\begin{table}[!t]
  \centering
  \resizebox{1\linewidth}{!}{%
    \begin{tabular}{crrrrrr}
    \toprule
    Model & \multicolumn{6}{c}{QwenVL2.5 7B} \\
    \midrule
    Casual Types & \multicolumn{1}{c}{ACC} & \multicolumn{1}{c}{ACC-R} & \multicolumn{1}{c}{ACC Group} & \multicolumn{1}{c}{F1 Score} & \multicolumn{1}{c}{R-Score} & \multicolumn{1}{c}{I-Score} \\
    \midrule
    CC    & 25.00 & 85.42 & 20.83 & 50.71 & 2.67  & 2.60 \\
    EM    & 50.76 & 68.94 & 36.36 & 59.51 & 2.72  & 3.31 \\
    AP    & 62.50 & 100.00 & 62.50 & 80.57 & 2.63  & 2.50 \\
    NP    & 37.61 & 57.26 & 18.80 & 46.92 & 1.32  & 0.79 \\
    PC    & 35.56 & 93.33 & 28.89 & 61.21 & 2.64  & 2.84 \\
    \bottomrule
    \end{tabular}%
    }
    \caption{Classification performance of the QwenVL2.5 7B, which achieves the SOTA performance among the Open-Source VLMs in our TimeCasualty.}
    \vspace{-0.5cm}
  \label{tab:sota_opensource}%
\end{table}%



\section{Conclusion}

In this work, we introduce \textbf{TimeCasualty}, a novel benchmark aimed at evaluating the temporal-causal reasoning capabilities of VLMs with a three-aspect evaluation pipeline. The result shows that, despite the current VLMs' impressive achievements in various tasks, they face notable limitations in comprehending deeper causal and commonsense knowledge. With the high-quality data and the comprehensive evaluation, TimeCasualty highlights the shortcomings in the causal reasoning ability of VLMs, which are common but are always ignored in the current evaluation.

\section{Limitations}
While TimeCausality evaluates how current VLMs understand temporal causality, it is subject to certain limitations: (i) The scope of temporal causality is extensive. Although we propose a classification, developing a benchmark that comprehensively covers all facets of these causal relationships is inherently challenging. (ii) Our work introduces a benchmark and reveals that existing VLMs often demonstrate suboptimal performance on it; however, it is still difficult to address these identified weaknesses. Investigating methods to enhance the capabilities of VLMs in analyzing such real-world causal relationships constitutes an important avenue for future research.

\bibliography{main}
\appendix
\onecolumn

\begin{center}
    \LARGE \textbf{Appendix For ``TimeCausality: Evaluating the Causal Ability in Time Dimension for
Vision Language Models''}
\end{center}
\vspace{1em} 

\section{Experiments Details.}\label{sec:exp_deatails}
For the Closed-Source VLMs, we revoke the official api for evaluation. For the Open-Source VLMs, we use their corresponding weight on Huggingface\footnote {https://huggingface.co/}, with the package\footnote{https://github.com/InternLM/lmdeploy} to deploy them locally with the default setting. All locally deployed experiments are conducted on a server with NVIDIA RTX4090s.

\section{More Cases in TimeCausality}
\label{sec:more_cases_in_time}
\begin{figure}[H]
  \centering
   \includegraphics[width=\linewidth]{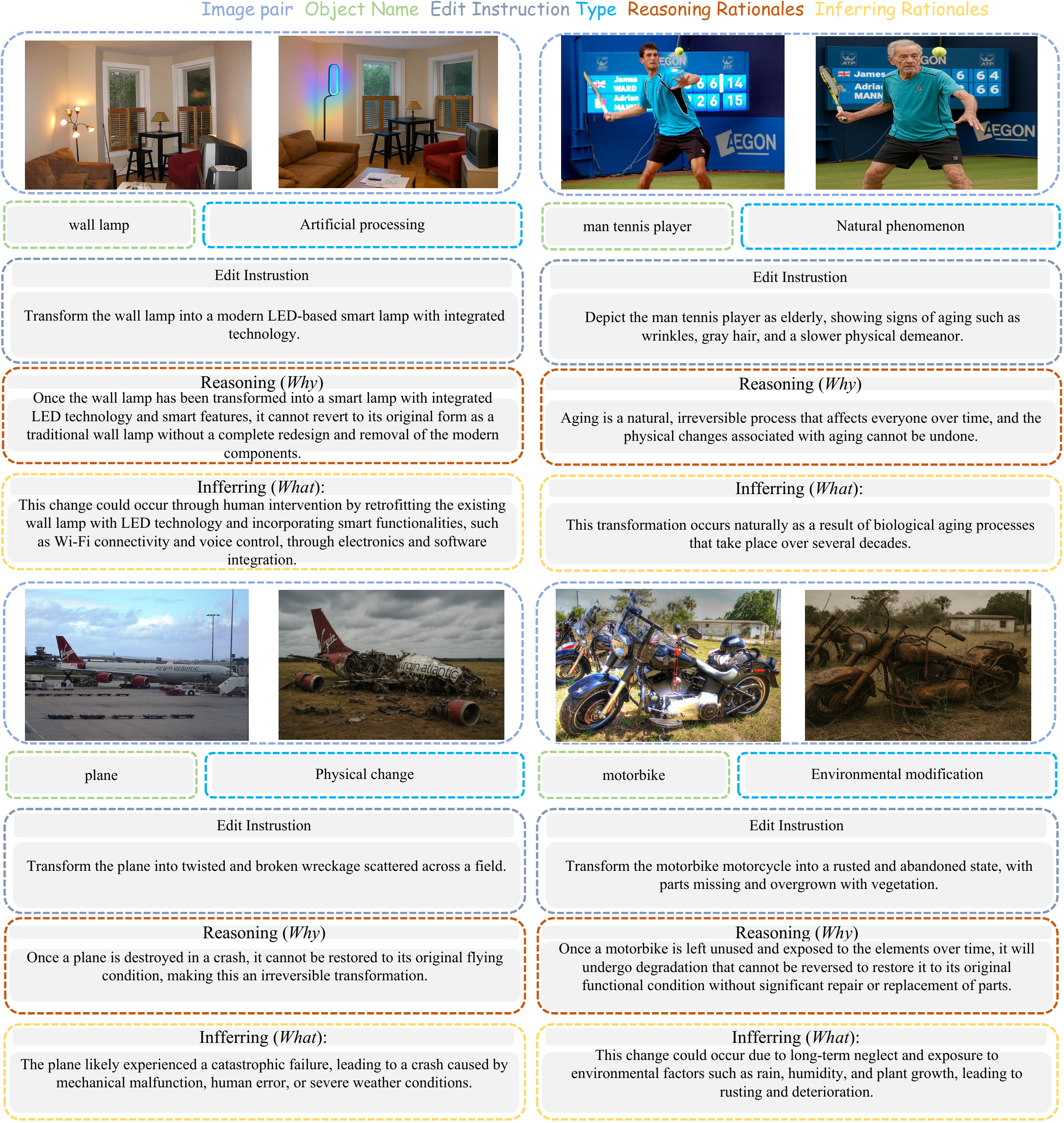}
   \caption{More Cases in Our Time Causality}
   \label{fig:more_cases_in_time}
\end{figure}

\setcounter{page}{1}
\setcounter{figure}{0}
\renewcommand\thefigure{A\arabic{figure}}
\clearpage

\section{The Gap Between Aspect I with II\&III}
\label{sec:a_sepcial_case}
\begin{figure}[H]
  \centering
   \includegraphics[width=0.7\linewidth]{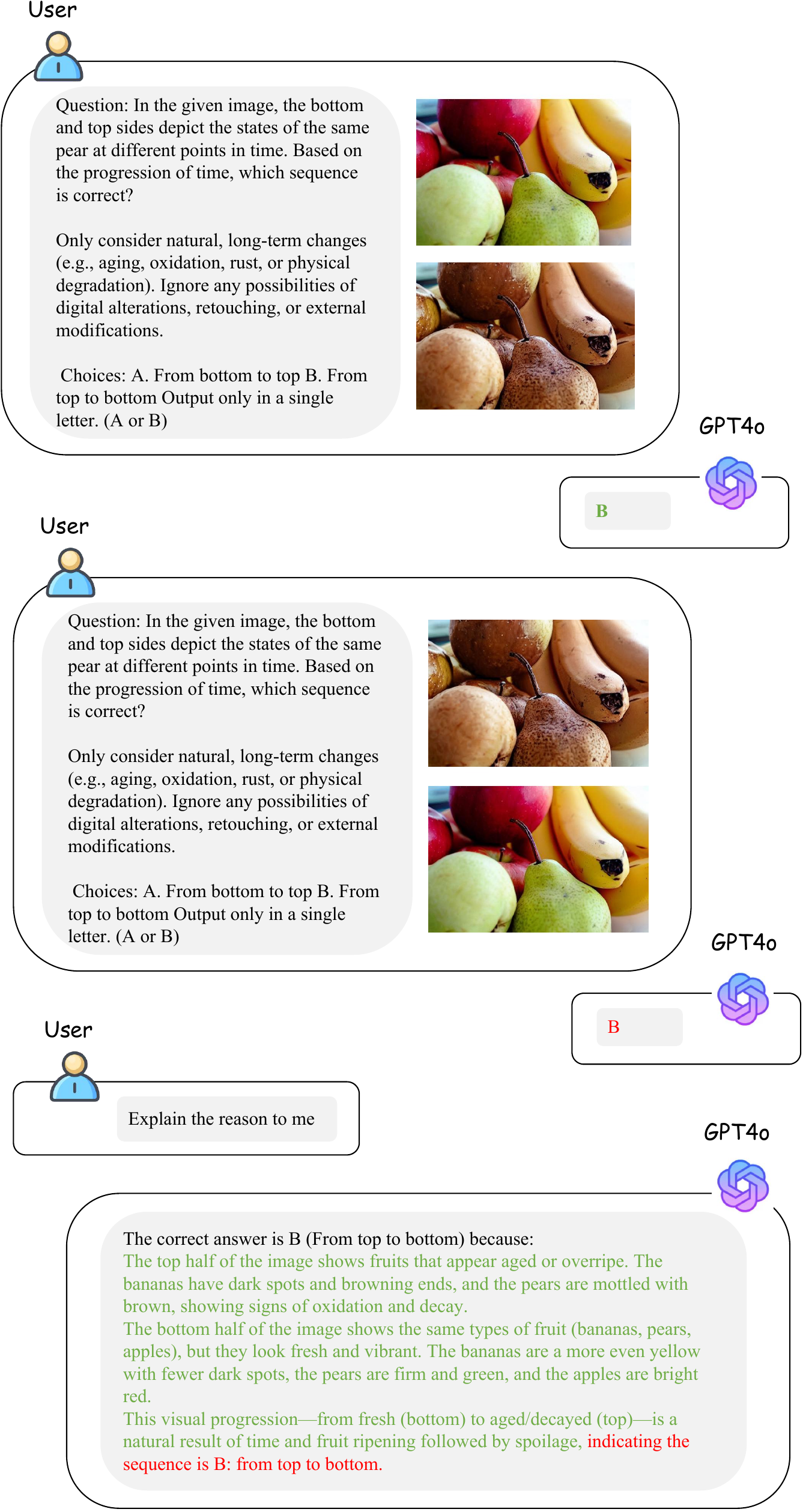}
   \caption{What makes VLMs fail to solve our proposed task? In this case, GPT4o successfully identifies the temporal order of in the first image-pair, but fails to do so on the reversed image. This shows an important point in the Aspect I evaluation: consistency. At the same time, for the reversed image-pair, GPT4o gives a completely correct reasoning process, but gives a wrong answer. This shows that although VLM may have certain capabilities in Aspect II\&III, it does not successfully associate them with visual information, resulting in the wrong final answer.}
   \label{fig:a_special_case}
\end{figure}

\clearpage
\section{Prompts}
\label{sec:appendix_prompts}
\subsection{Prompt and Models for Dataset Generation}
\label{subsec:prompt_generation}
We use the Grounded-Segment-Anything\footnote{https://github.com/IDEA-Research/Grounded-Segment-Anything} as the Auto-Label Model, and the GPT4o as the LLM and the inpainting model.
\begin{table}[H]
\begin{center}
\begin{tcolorbox}[
  colframe=blue!20!black,    
  colback=gray!10,           
  coltitle=black,              
  colbacktitle=blue!20,        
  title={Example prompt for data generation}
]

Based on the provided object describe an irreversible transformation it could undergo. The transformation should fit one of the following categories:\\

1.Physical change: Structural or morphological changes caused by external forces or environmental factors (e.g., a glass cup shattered).\\
2.Chemical change: Changes to the object’s chemical composition, resulting in new substances (e.g., food rotting).\\
3.Natural phenomenon: Changes due to natural processes or the passage of time (e.g., aging of a person).\\
4.Environmental modification: Alterations caused by human intervention over time (e.g., a historical building renovated into a modern structure).\\
5.Artificial processing: Transformations caused by human-made techniques or craftsmanship (e.g., raw clay turned into pottery).\\

For each input object:\\
• Provide a clear edit instruction for the irreversible transformation.\\
• Justify why this transformation logically follows from the original state and cannot revert.\\
• Explain how this change could occur, referencing plausible causes or processes.\\
• Specify the type of the transformation.\\
Output your response in the following structured format:\\
<type>\\

[Specify the type of change, e.g., "Physical change", "Chemical change", "Natural phenomenon", "Environmental modification", or "Artificial processing."]\\

</type>\\

<edit\_instruction>\\

[Provide a concise and clear directive for the transformation, e.g., “Transform the glass cup into shattered pieces scattered across a surface.”]\\

</edit\_instruction>\\

<rationales>\\

[Explain why the modified image depicts a later state in time or process compared to the original, e.g., "A shattered glass cannot return to its original intact state, making this an irreversible transformation."]\\

</rationales>
\label{example_prompt}
\end{tcolorbox}
\end{center}
\end{table}

\begin{table}[H]
\centering
\begin{tcolorbox}  

<operation>\\

[Describe what could cause the change to occur, e.g., "The glass was likely subjected to a sudden external force or impact."]\\

</operation>\\

Examples:\\

Input: A glass cup\\
<type>Physical change</type>\\
<edit\_instruction> Transform the glass cup into shattered pieces scattered across a surface. </edit\_instruction>\\
<rationales>A shattered glass cannot return to its original intact state, making this an irreversible transformation.</rationales>\\
<operation>The glass was likely subjected to a sudden external force or impact.</operation>\\

Input: A slice of bread\\
<type>Chemical change</type>\\
<edit\_instruction> Show the bread as moldy and green with visible spores on its surface. </edit\_instruction>\\
<rationales>Once bread molds, the chemical composition changes, and it cannot revert to its original edible state.</rationales>\\
<operation>The bread likely underwent chemical reactions caused by exposure to moisture and microbial activity over time.</operation>\\

Input: A person\\
<type>Natural phenomenon</type>\\
<edit\_instruction> Depict the person as elderly, with wrinkles, gray hair, and visible signs of aging. </edit\_instruction>\\
<rationales>Aging is a natural, irreversible process that occurs over time and cannot be reversed to a previous state.</rationales>
<operation>The changes occur naturally due to biological aging processes over years.</operation>\\

Input: A small village\\
<type>Environmental modification</type>\\
<edit\_instruction> Transform the small village into a modern urban area with tall buildings and paved roads. </edit\_instruction>\\
<rationales>Urbanization is an irreversible process that permanently alters the original layout and structure of the village.</rationales>
<operation>This change could occur due to long-term urban development and modernization initiatives by humans.</operation>\\

Now, for the following input object, apply the same structure:\\

Input: \texttt{\{Input Source\}}\\

\end{tcolorbox}
\label{tab:example_prompt_generation}
\end{table}

\subsection{Prompt for LLM-as-a-Judge}
\label{sec:prompt_judge}

\begin{table}[H]
\begin{center}
\begin{tcolorbox}[
  colframe=blue!20!black,    
  colback=gray!10,           
  coltitle=black,              
  colbacktitle=blue!20,        
  title={Example prompt for LLM-as-a-Judge}
]
You will be given a ground truth and model output couple.\\
Your task is to provide a 'total rating' scoring how well the model output matches the semantic meaning of the ground truth.\\
Give your answer as an integer on a scale of 0 to 5, where 0 means that the model output is completely unrelated to the ground truth, and 5 means that the model output perfectly matches the semantic meaning of the ground truth.\\

Provide your feedback as follows:\\

Feedback:\\
Total rating: (your rating, as an integer between 0 and 5)\\

Now here are the ground truth and model output.\\

Ground Truth: \texttt{\{ground\_truth\}}\\
Model Output: \texttt{\{model\_output\}}\\

Feedback:\\
Total rating: \\

\label{tab:example_prompt_judge}
\end{tcolorbox}
\end{center}
\end{table}

\subsection{Prompt for Evaluation}
\label{sec:prompt_evaluation}
\begin{table}[H]
\begin{center}
\begin{tcolorbox}[
  colframe=blue!20!black,    
  colback=gray!10,           
  coltitle=black,              
  colbacktitle=blue!20,        
  title={Example prompt for Aspect I Evaluation}
]

Question: In the given image, the bottom and top sides depict the states of the same \texttt{\{Source Object\}}  at different points in time. Based on the progression of time, which sequence is correct?\\
Only consider natural, long-term changes (e.g., aging, oxidation, rust, or physical degradation). Ignore any possibilities of digital alterations, retouching, or external modifications.\\
Choices:\\
A. From bottom to top (Bottom happens first, top happens later). \\
B. From top to bottom (Top happens first, bottom happens later). \\
Output only in a single letter. (A or B) .\\

\label{tab:example_prompt_evaluation}
\end{tcolorbox}
\end{center}
\end{table}

\begin{table}[H]
\begin{center}
\begin{tcolorbox}[
  colframe=blue!20!black,    
  colback=gray!10,           
  coltitle=black,              
  colbacktitle=blue!20,        
  title={Example prompt for Aspect II \& III Evaluation}
]

Why does the \texttt{\{Source Object\}} in the scene on the top occur after the one on the bottom? \\
Answer in one sentence.\\
\\
What caused the transition from the \texttt{\{Source Object\}} on the bottom to the \texttt{\{Source Object\}} on the top? Answer in one sentence.

\label{tab:example_prompt_evaluation}
\end{tcolorbox}
\end{center}
\end{table}

\begin{table*}[H]
\begin{center}
\begin{tcolorbox}[
  colframe=blue!20!black,    
  colback=gray!10,           
  coltitle=black,              
  colbacktitle=blue!20,        
  title={Example prompt for Aspect III Evaluation}
]

What caused the transition from the \texttt{\{Source Object\}} on the bottom to the \texttt{\{Source Object\}}  on the top? \\
Answer in one sentence.\\

\label{tab:example_prompt_evaluation}
\end{tcolorbox}
\end{center}
\end{table*}

\clearpage
\section{Human Verification.}\label{sec:human_verification}

\begin{table*}[h]
\begin{center}
\begin{tcolorbox}[      
  title={Guidelines for Image Pair Annotation: }
]

This is an image pair showing two states of the same scene or object. Please evaluate the following: \\

Temporal Order: Do these two images have a clear sequential relationship? In other words, does the content of one image obviously appear before or after the other?\\

Content Consistency: Do the two images maintain overall consistency? Can you confirm that they depict the same scene or the same object?\\

Reasoning Rationales: Does the reason for the order of these two images align with the provided {Reasoning Rationales}?\\

Inferring Rationales: Does the reason for the order of these two images align with the provided {Inferring Rationales}?\\

Final Check: Please also assess whether this image pair contains any disturbing or NSFW (Not Safe For Work) content, such as horror, gore, or other inappropriate material.\\

\label{tab:example_prompt_evaluation}
\end{tcolorbox}
\end{center}
\end{table*}

\end{document}